\pdfoutput=1

\documentclass[11pt]{article}

\usepackage[final]{acl}

\usepackage[most]{tcolorbox} 

\usepackage[utf8]{inputenc}
\usepackage{multirow}
\usepackage{booktabs}
\usepackage{geometry}
\usepackage{pgfplots} 
\pgfplotsset{compat=1.17} 
\usepackage{tikz} 

\usepackage{times}
\usepackage{latexsym}

\usepackage[T1]{fontenc}

\usepackage[utf8]{inputenc}

\usepackage{microtype}

\usepackage{inconsolata}

\usepackage{graphicx}

\title{Adaptive Linguistic Prompting (ALP) Enhances Phishing Webpage Detection in Multimodal Large Language Models}
\author{%
Atharva Bhargude\thanks{Lead Author} 
\quad Ishan Gonehal 
\quad Dave Yoon \\
\quad \textbf{Kaustubh Vinnakota} 
\quad \textbf{Chandler Haney}
\quad \textbf{Aaron Sandoval}\thanks{Senior Author} 
\quad \textbf{Kevin Zhu}\footnotemark[2] 
\\
Algoverse AI Research\\
\texttt{atharvab167@gmail.com, kevin@algoverseacademy.com}
}

\begin{document}
\maketitle
\begin{abstract}

Phishing attacks represent a significant cybersecurity threat, necessitating adaptive detection techniques. This study explores few-shot Adaptive Linguistic Prompting (ALP) in detecting phishing webpages through the multimodal capabilities of state-of-the-art large language models (LLMs) such as GPT-4o and Gemini 1.5 Pro. ALP is a structured semantic reasoning method that guides LLMs to analyze textual deception by breaking down linguistic patterns, detecting urgency cues, and identifying manipulative diction commonly found in phishing content. By integrating textual, visual, and URL-based analysis, we propose a unified model capable of identifying sophisticated phishing attempts. Our experiments demonstrate that ALP significantly enhances phishing detection accuracy by guiding LLMs through structured reasoning and contextual analysis. The findings highlight the potential of ALP-integrated multimodal LLMs to advance phishing detection frameworks, achieving an F1-score of 0.93—surpassing traditional approaches. These results establish a foundation for more robust, interpretable, and adaptive linguistic-based phishing detection systems using LLMs.
\end{abstract}

\section{Introduction}

With over 1.2 million attempts blocked in 2024 alone \cite{acharya2024}, phishing remains a persistent cybersecurity threat as attackers continuously refine their tactics to evade detection \cite{10735206}. Traditional detection methods, such as heuristic URL matching and brand verification, often falter against novel evasion tactics \cite{li2024knowphishlargelanguagemodels}. While earlier work \cite{lee2024multimodallargelanguagemodels} compared modality contributions using zero-shot brand + domain verification, we instead focus on how prompt engineering alone can enhance phishing detection performance using Multimodal Large Language Models (MLLMs).

\indent The emergence of Large Language Models (LLMs) and Multimodal Large Language Models (MLLMs), represents a paradigm shift in phishing detection \cite{SecureList2023}, offering fine-grained semantic analysis. These MLLMs, including GPT-4o and Gemini 1.5 Pro, enable joint reasoning over HTML, screenshots, and URLs \cite{touvron2023llamaopenefficientfoundation}. 

\indent In this paper, we introduce \textbf{Adaptive Linguistic Prompting (ALP)}, a method inspired by few-shot prompting (\citeauthor{brown2020languagemodelsfewshotlearners}, \citeyear{brown2020languagemodelsfewshotlearners}; \citeauthor{agrawal2022largelanguagemodelsfewshot}, \citeyear{agrawal2022largelanguagemodelsfewshot}) and structured reasoning \cite{wei2023chainofthoughtpromptingelicitsreasoning}. It is a few-shot prompting framework that guides LLMs to perform structured, modality-specific reasoning using curated exemplars rather than zero-shot templates. Our contributions include: (1) a refined 8-shot prompting framework (Prompt-Enhanced ALP) for HTML + Screenshot and URL analysis, and (2) design insights—such as a “suspicious-first” URL heuristic—that improve F1 (0.91 vs.\ 0.93), demonstrating the impact of prompt tuning independent of model changes.

\section{Related Works}
Traditional phishing detection relies on heuristics like URL analysis, HTML structure analysis, and blacklisting \cite{li2024knowphishlargelanguagemodels}. While effective for known threats, these methods fail against zero-day attacks and sophisticated mimicry tactics \cite{kulkarni2024mlllmevaluatingrobustness}. Machine learning (ML)-based models extract statistical patterns to improve detection (\citeauthor{whittaker2010large}, \citeyear{whittaker2010large}; \citeauthor{article}, \citeyear{article}), but adversarial perturbations and dynamic content limit their effectiveness \cite{lee2024multimodallargelanguagemodels}.

Recent phishing detection methods use computer vision to spot brand imitation via logos and page layout
(\citeauthor{abdelnabi2020visualphishnet}, \citeyear{abdelnabi2020visualphishnet}; \citeauthor{lin2021phishpedia}, \citeyear{lin2021phishpedia}; \citeauthor{ji2024evaluatingeffectivenessrobustnessvisual} \citeyear{ji2024evaluatingeffectivenessrobustnessvisual}), however these models require continuous retraining.

\citeauthor{lee2024multimodallargelanguagemodels}, \citeyear{lee2024multimodallargelanguagemodels} proposed a multimodal LLM approach integrating brand identification and domain verification to detect phishing inconsistencies. Their findings show LLMs outperform traditional methods in accuracy and robustness but face challenges like prompt injection and high computational costs (\citeauthor{divakaran2024llmscybersecuritynew} \citeyear{divakaran2024llmscybersecuritynew}). Other models, like ChatSpamDetector (\citeauthor{koide2024chatspamdetector}, \citeyear{koide2024chatspamdetector}), also leverage LLMs. Our research advances this field by integrating few-shot ALP prompting to refine linguistic feature extraction and reasoning, reducing dependence on continuous retraining while strengthening phishing detection.
    
\section{Methodology}

        For the experiments, we selected the GPT-4o and the Gemini 1.5 Pro models for their advanced multi-modal analysis architectures, both of which effectively correlate verbose textual content and visual inputs \cite{Yin_2024}. This is crucial for phishing detection, where identifying malicious sites often requires analyzing a combination of HTML content, syntactic patterns, and screenshots.

\subsection{Data Collection}
We use the curated \citet{lee2024multimodallargelanguagemodels} dataset sourced from \citet{Lee2024GitHub}, containing 1607 Benign Brands and 289 Phishing Brands; each brand is represented by one screenshot and one JSON file containing HTML attributes. 311/1607 Benign Brands and all 289 Phishing Brands are retained through our filtering approach, to optimize computational efficiency.
\begin{figure}
  \centering
  \includegraphics[width=0.32\textwidth]{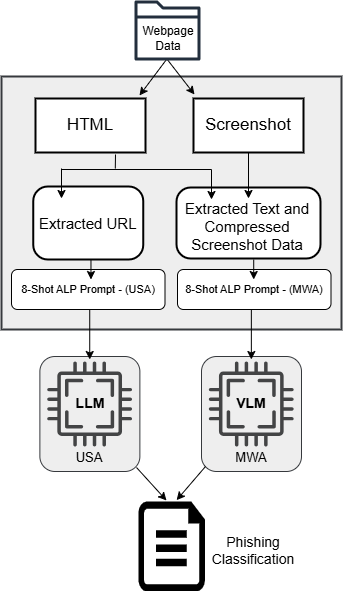}
  \caption{Overview of the Multimodal VLM \& LLM Phishing Webpage Detection System.}
  \label{fig:experiments}
\end{figure}
\subsection{Dataset Filtering}
To optimize computational efficiency while preserving dataset diversity, the benign dataset is curated to 311 brands. This size mitigates token-based processing costs in GPT-4o while preserving analytical rigor. Our filtering approach (automated clustering paired with manual validation) prioritizes domain diversity, incorporating both high-reputation entities (e.g., "google.com", "amazon.com") and traffic-tiered domains identified through web-traffic metrics by \cite{ExplodingTopics2024}. Each brand is represented by two standardized instances—one screenshot and one JSON file containing HTML attributes—to streamline processing while capturing multimodal features critical for phishing detection. The selection criteria emphasizes minimizing redundancy without compromising domain heterogeneity, ensuring coverage of both commonly accessed and niche pages.

\subsection{Few-Shot ALP Prompting Methodology}
This paper introduces a novel prompting methodology that integrates few-shot Adaptive Linguistic Prompting (ALP) to enhance phishing detection using LLMs for the analysis of multimodal data, including text, images, and URLs.
\indent We build on \citeauthor{lee2024multimodallargelanguagemodels}'s data splits and baseline prompts, reimplementing their zero-shot brand and domain setup, refining 8-shot prompts for HTML + Screenshot and URL analysis using held-out validation, and evaluating all methods on F1, precision, and recall.

Two distinct analyses were implemented to leverage the multi-modal capabilities of GPT-4o and Gemini 1.5 Pro for phishing detection. Both analyses utilized the curated dataset, ensuring consistent input modalities. 

The first analysis, Multimodal Webpage Analysis (MWA), utilizes multi-modal capabilities by analyzing both the HTML textual content and the screenshot data of a given webpage. The diagram (\autoref{fig:experiments}) illustrates the flow of MWA prompting, wherein HTML and screenshot data are extracted and processed through an 8-shot ALP reasoning prompt.   \\
{1. Multimodal Webpage Analysis (MWA) Instruction Prompt utilizing ALP}
\begin{tcolorbox}[
    colback=gray!10,
    colframe=gray!50,
    sharp corners,
    boxrule=0.5pt,
    left=5pt,
    right=5pt,
    top=5pt,
    bottom=5pt,
    width=\linewidth, 
    boxsep=2pt 
]
You are a phishing detection assistant. Given a randomly selected phishing or benign URL, you will analyze its HTML and screenshot data. Use the provided few-shot examples as guides to assess phishing risk. Structure your analysis into brand recognition, diction and syntax, and screenshot-specific evaluations.  \\

(HTML and screenshot input)\\
(8-shot ALP examples)\\

Your response should follow this order:

    1. Step-by-Step Reasoning: 
    
    2. Key Indicators:
    
    3. Supporting Evidence: 
    
    4. Confidence Score 0-10: 
    
    5. Phishing Risk (Phishing/Benign): 
\end{tcolorbox}

The second analysis, URL Structure Analysis (USA), applies URL-matching techniques from existing literature, leveraging structured data from embedded URLs to classify phishing attempts. In this analysis, zipped files containing website data are processed systematically, with each HTML file being parsed to extract URLs. These URLS are then processed through an 8-shot ALP reasoning prompt and are then used to evaluate the model's ability to identify phishing through URL analysis.\\\\
{2. URL Structure Analysis (USA) Instruction Prompt}
\begin{tcolorbox}[
    colback=gray!10,
    colframe=gray!50,
    sharp corners,
    boxrule=0.5pt,
    left=5pt,
    right=5pt,
    top=5pt,
    bottom=5pt,
    width=\linewidth, 
    boxsep=2pt 
]
You are a phishing detection expert. Your task is to classify URLs as only either 'phishing' or 'benign' and provide a detailed explanation of why. Focus on key features like domain name, protocol, URL path, and potential phishing indicators.\\

(URL input)\\
(8-shot MWA examples)\\

Your response should follow this order:

URL:

Features:

Reasoning:

Label: (Benign or Phishing)
\\
\end{tcolorbox} 
\subsection{Combining Analysis Results} 
We fuse MWA and USA outputs with a straightforward, risk‑aware rule. When both analyses agree, we accept that label immediately. If they disagree, we label phishing whenever USA predicts it or when MWA’s confidence exceeds 8.5—otherwise we label benign. The 8.5 cutoff was set a priori on a training split to balance precision and recall.\\
USA, optimized for URL analysis, is adept at identifying domain-specific anomalies, while MWA is adept at capturing subtler phishing signals, such as content and visual mimicry. The combined analysis enhances detection accuracy, particularly in cases involving visually deceptive phishing tactics.
\begin{table*} 
    \centering
    \begin{tabular}{lcccc}  
    \hline
    Model & Approach & Precision & Recall & F1 \\
    \hline
    \multirow{4}{*}{GPT-4o} 
    & Baseline & 0.91 & 0.91 & 0.91 \\
    & MWA & 0.80 & 0.89 & 0.84 \\
    & USA & 0.91 & 0.91 & 0.91 \\
    & Combined Analysis & 0.91 & 0.94 & 0.93 \\
    \hline
    \multirow{4}{*}{Gemini 1.5 Pro}
    & Baseline & 0.76 & 0.85 & 0.81 \\
    & MWA & 0.94 & 0.87 & 0.90 \\
    & USA & 0.88 & 0.85 & 0.87 \\
    & Combined Analysis & 0.91 & 0.92 & 0.91 \\
    \hline
    \end{tabular}
    \caption{Performance comparison across baseline, MWA, USA, \& Combined Analysis}
    \label{tab:performance}
\end{table*}

\begin{figure}[ht]
    \centering
    \centering
\begin{minipage}[t]{0.45\textwidth}
        \centering
        \begin{tikzpicture}
        \begin{axis}[
            width=\textwidth, 
            height=5.5cm, 
            ybar,
            bar width=8pt, 
            ylabel={F1-score},
            symbolic x coords={Gemini 1.5 Pro, GPT-4o},
            xtick=data,
            enlarge x limits=0.6,
            ymin=0.75,
            ymax=1,
            ytick={0.75,0.80,0.85,0.90,0.95,1}, 
            grid=major,
            major grid style={line width=0.5pt,draw=gray!50},
            legend style={at={(0.5,-0.2)},anchor=north,legend columns=-1},
            legend entries={Baseline, MWA, USA, Combined},
        ]
        \addplot coordinates {(Gemini 1.5 Pro, 0.81) (GPT-4o, 0.91)};
        \addplot coordinates {(Gemini 1.5 Pro, 0.90) (GPT-4o, 0.84)};
        \addplot coordinates {(Gemini 1.5 Pro, 0.87) (GPT-4o, 0.91)};
        \addplot coordinates {(Gemini 1.5 Pro, 0.91) (GPT-4o, 0.93)};
        \end{axis}
        \end{tikzpicture}
        \caption{F1-score for Gemini 1.5 Pro and GPT-4o between Baseline, MWA, USA, \& Combined Analysis}
        \label{fig:F1}
    \end{minipage}

\end{figure}
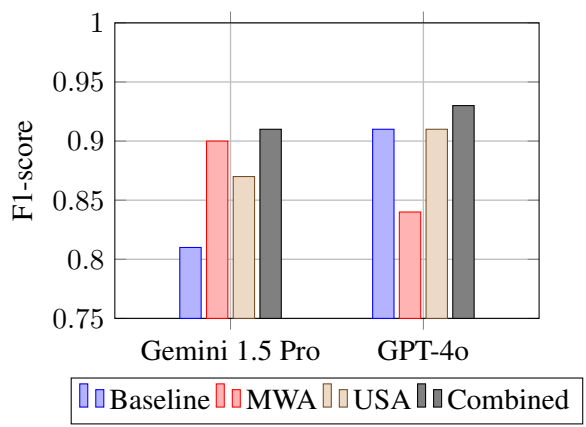

\section{Results and Discussion}

The results demonstrate the efficacy of multimodal phishing detection modified by Adaptive Linguistic Prompting (ALP) over heuristic zero-shot prompting methods seen in our baseline \cite{lee2024multimodallargelanguagemodels}.

The performance assessment revealed different strengths between GPT-4o and Gemini 1.5 Pro across MWA, USA, and the combined analysis. MWA outperformed static approaches by detecting advanced deception tactics, particularly brand mimicking. Gemini 1.5 Pro performed well in identifying brand representation anomalies due to its comprehensive analysis of visual elements and textual consistency markers.

USA demonstrated complementary strengths, with GPT-4o excelling in domain authenticity and URL pattern analysis, identifying suspicious structures and security protocol inconsistencies. The combined analysis improved performance by integrating both approaches.

GPT-4o's combined analysis achieved a precision of 91.67\%, recall of 94.12\%, and an F1-score of 0.93, while Gemini 1.5 Pro achieved 91\% precision, 92\% recall, and an F1-score of 0.91. These results suggest an optimized phishing detection framework could integrate both models, with GPT-4o for USA and Gemini for MWA to maximize accuracy.

While testing USA, we incorporated a "suspicious-first" prompting strategy into USA, as seen below.
\begin{tcolorbox}[
    colback=gray!10,
    colframe=gray!50,
    sharp corners,
    boxrule=0.5pt,
    left=5pt,
    right=5pt,
    top=5pt,
    bottom=5pt,
    width=\linewidth, 
    boxsep=2pt 
]
**If you are unsure and feel that a link is suspicious of phishing activity, label it phishing**

**Otherwise, if the link relates to an official domain, label it Benign**

\end{tcolorbox}
 This guided models to classify uncertain cases as potential threats, significantly improving USA detection accuracy while maintaining a practical balance between security and usability. This finding suggests prompt engineering could be as crucial as model architecture in developing robust detection systems as such a simple prompt increased USA's accuracy from 81\% to 91\% in GPT-4o by adopting a risk-averse approach.
\section{Conclusion}
Our study demonstrates that Adaptive Linguistic Prompting (ALP) enhances phishing detection by guiding multimodal LLMs to systematically analyze brand, linguistic, visual, and URL-based deception tactics. By integrating semantic reasoning and few-shot prompting, ALP addresses critical gaps in traditional heuristic methods, which struggle with nuanced phishing cues (examples in \hyperref[sec:appen]{Appendix A.1}). The combined analysis function achieved an F1-score of 0.93, a significant improvement over the baseline and past literature, highlighting ALP’s ability to harmonize detection modalities while prioritizing risk-averse classification. Our findings establish ALP as a robust, interpretable, and scalable solution, reducing reliance on continuous retraining. 
\section{Limitations}
With frameworks utilizing ALP to advance phishing detection, certain limitations merit acknowledgment. First, the dataset, though diverse, may not fully capture emerging phishing tactics or region-specific attacks, potentially affecting applicability. Second, reliance on proprietary LLMs like GPT-4o introduces scalability and cost barriers, limiting accessibility for broader deployment. Third, the framework’s performance on non-English content and adversarial evasion strategies (e.g., context-aware paraphrasing) remains unexplored. Fourth, the "suspicious-first" strategy prioritizes URL analysis in conflicting MWA and USA situations, which could marginalize nuanced visual mimicry detected by multimodal analysis. A weighted probability combination function could potentially be explored. Finally, comparisons to machine learning architectures other than transformer-based LLMs are absent. Future research should optimize prompt engineering to lower token consumption and explore multi-LLM frameworks to leverage specific model strengths. Further, expanding datasets to cover non‑English phishing tactics, integrating cost‑effective open‑source models, and rigorously testing adversarial evasion strategies could strengthen ALP and enhance phishing detection accuracy.




\bibliography{latex/custom}

\onecolumn
\appendix
\section{Appendix}
\label{sec:appen}
\subsection{Few-Shot-Prompts}
Below is a few-shot prompt for MWA
\begin{tcolorbox}[
    colback=gray!10,
    colframe=gray!50,
    sharp corners,
    boxrule=0.5pt,
    left=5pt,
    right=5pt,
    top=5pt,
    bottom=5pt,
    width=\linewidth, 
    boxsep=2pt 
]
Example 1 \\
Url: https://secure-appleid-login.com/\\
HTML Data: title: Apple ID Login meta description: Sign in to your
Apple ID account. favicon: https://www.apple.com/favicon.ico logo alt text:
Apple Logo footer text: Not Found headers text: Home | Support | Account |
Privacy nav bar content: Not Found paragraphs text: Enter your Apple ID
and password to sign in. span text: Forgot Apple ID? | Forgot Password? |
Sign Up\\
Screenshot Info: The Apple logo is high-resolution and matches official
branding. The login form includes fields for Apple ID and password. The
"Sign In" button is styled with Apple's official colors. The footer lacks
additional information and links typically found on Apple's official site.
Brand Recognition Analysis: The page uses the official Apple logo
with correct alt text. The URL ("secure-appleid-login.com") is not an official
Apple domain. HTTPS protocol is absent, which is unusual for Apple’s
official login pages.\\
Diction and Syntax Analysis: The language is straightforward without
excessive capitalization. Instructions are clear and mimic official Apple
communications. There is no urgent or threatening language detected.
Screenshot Analysis: Visual elements largely mimic Apple’s official login
page. However, the absence of additional footer links and the domain
discrepancy raise suspicion.\\
Step-by-Step Reasoning:\\
1. Branding elements appear legitimate with a correct Apple logo.\\
2. The URL does not match Apple’s official domain and lacks HTTPS.\\
3. Missing footer elements typical of Apple's official site suggest
phishing.\\
Classification: PHISHING
\end{tcolorbox}

The Multimodal Webpage Analysis (MWA) few-shot prompt guides the model to analyze phishing risk by breaking down the evaluation into structured components: brand recognition, diction and syntax, and screenshot-specific analysis. It provides step-by-step reasoning, key indicators, supporting evidence, a confidence score, and a final phishing/benign classification. This approach ensures the model systematically analyzes both textual and visual elements, leveraging examples to identify subtle inconsistencies in branding, language, and design that may indicate phishing.\\

Below is a few-shot prompt for USA
\begin{tcolorbox}[
    colback=gray!10,
    colframe=gray!50,
    sharp corners,
    boxrule=0.5pt,
    left=5pt,
    right=5pt,
    top=5pt,
    bottom=5pt,
    width=\linewidth, 
    boxsep=2pt 
]
Example 1:\\
Link: 'http://university-scholarship-app.us/confirm-details'\\
Features: The domain includes terms such as "university" and "scholarship", HTTP protocol, path "confirm-details”.\\
Reasoning: This link exhibits several traits of a phishing attempt. The domain uses generic terms like "university" and "scholarship," which are unrelated to any well-known institution, making it highly suspicious. The use of HTTP instead of HTTPS suggests that the site does not encrypt user data, leaving it vulnerable to interception. The path, "confirm-details," is a classic phishing tactic designed to solicit sensitive personal or account-related information under the guise of legitimacy. These factors combined strongly indicate malicious intent.\\
Classification: PHISHING
\end{tcolorbox}
The URL Structure Analysis (USA) few-shot prompt focuses on classifying URLs as phishing or benign based on domain name, protocol, URL path, and phishing indicators. It requires the model to provide detailed reasoning for its classification, emphasizing features like HTTPS usage, domain legitimacy, and suspicious path structures. By incorporating examples of phishing URLs, the prompt trains the model to recognize patterns such as generic domain names, insecure protocols, and deceptive paths, enabling it to detect phishing attempts through URL analysis alone.\\

\subsection{Model Outputs}
The following focuses on four different model outputs over MWA and USA for GPT and Gemini. The webpage in focus is '\url{https://www.12voltdoesit.com/}'

\textit{1. GPT-4o MWA Response}

\begin{tcolorbox}[
    colback=gray!10,
    colframe=gray!50,
    sharp corners,
    boxrule=0.5pt,
    left=5pt,
    right=5pt,
    top=5pt,
    bottom=5pt,
    width=\linewidth, 
    boxsep=2pt 
]
Folder Name: 12voltdoesit
AI Response:
1. *Step-by-Step Reasoning*: 
   - **Brand Recognition Analysis**: The HTML content mentions the website title as "Home | 12VOLTDOESIT - Automotive Electronics Services," indicating a focus on automotive electronics. However, the favicon and logo alt text are marked as "Not Found," which raises suspicion about the legitimacy of the brand. The footer text provides a service description but lacks well-known brand indicators or established company credentials. The URL "12voltdoesit.com" does not correspond to a widely recognized brand.
   
   - **Diction and Syntax Analysis**: Phrases like "we would like to help" and "top-rated Automotive Electronic Services" suggest a friendly tone, but the overall structure seems cluttered with excessive product listings and pricing information, which could be seen as unprofessional. There is no urgent or alarming language, which is a positive sign.

   - **Screenshot Analysis**: The screenshot URL leads to an image that contains the visual representation of the webpage. The design is amateurish, potentially indicating a phishing attempt. The lack of identifiable branding elements in the HTML also reflects poorly in the screenshot.

2. *Phishing Risk (Phishing/Benign)*: Classification: Benign (Confidence score of 6.5 suggests some legitimacy, but caution is warranted due to missing brand recognition elements)

3. *Key Indicators*: 
   - Informal diction and promotional language
   - Cluttered presentation of services and products

4. *Confidence Score*: 6.50

5. *Supporting Evidence*: The analysis reveals that the absence of recognizable branding elements such as a logo or favicon raises red flags about its legitimacy. The informal nature of the text may indicate a lack of professionalism, which is common in phishing sites. However, the lack of urgent language and the coherent description of services suggest that it may not be an outright phishing attempt. The overall impression is that while there are indicators of concern, there is no definitive evidence to classify the site as phishing, leading to a benign classification.

6. *Provide URL*: URL: https://www.12voltdoesit.com/
\end{tcolorbox}

\textit{2. Gemini MWA Response}
\begin{tcolorbox}[
    colback=gray!10,
    colframe=gray!50,
    sharp corners,
    boxrule=0.5pt,
    left=5pt,
    right=5pt,
    top=5pt,
    bottom=5pt,
    width=\linewidth, 
    boxsep=2pt 
]
Folder Name: 12voltdoesit
AI Response:
1. *Step-by-Step Reasoning*: 
1. The URL seems legitimate, belonging to a business named "12voltdoesit" and using HTTPS.
2. Brand Recognition:  The branding is consistent throughout the website, though the lack of a logo image in the HTML is somewhat unusual for a business site. The footer copyright and contact information reinforce the legitimacy of the business.
3. Diction and Syntax Analysis: The language used on the site is generally clear and understandable, although there are some minor grammatical errors and repetitions, such as "Installation Include in price!". The language doesn't employ urgency or pressure tactics commonly found in phishing attempts.
4. Screenshot Analysis: The screenshot displays a professional-looking website with clear sections for services, products, and contact information. The screenshot matches the information provided in the HTML data, further reinforcing the website's authenticity.

2. *Phishing Risk*: Classification: BENIGN

3. *Key Indicators*: Repeated phrases "Installation Include in price!", some minor grammatical inconsistencies. Clear pricing for all advertised services. Presence of contact information, copyright information, and a physical address.

4. *Confidence Score*: 9.50

5. *Supporting Evidence*: The website exhibits several key characteristics of a legitimate business. It uses a proper domain name, secure HTTPS, and provides consistent branding across the HTML and the screenshot. While there are minor grammatical errors and repetitive phrases, these do not necessarily indicate a phishing attempt and could simply be due to oversight.  The absence of common phishing indicators like urgent language, requests for personal information (outside of standard login/registration), and misleading links further support the benign classification. The overall presentation and information provided create a high degree of confidence in its legitimacy.

6. *Provide URL*: URL: \url{https://www.12voltdoesit.com/}
\end{tcolorbox}

Regarding MWA comparisons, GPT-4o excels in identifying branding inconsistencies and linguistic informality, often flagging sites with missing favicons, unprofessional design, or cluttered content as suspicious. Its cautious approach leads to lower confidence scores (e.g., Confidence Score: 6.5) even when classifying sites as Benign, reflecting a focus on structural red flags. Gemini 1.5 Pro emphasizes branding coherence and visual authenticity, showing higher confidence in benign classifications (e.g., Confidence Score: 9.50) when branding elements align, even in the presence of minor linguistic errors. Its strength lies in recognizing legitimate visual and textual patterns, often overlooking subtle inconsistencies that GPT-4o flags.\\
{3. GPT-4o USA Response}
\begin{tcolorbox}[
    colback=gray!10,
    colframe=gray!50,
    sharp corners,
    boxrule=0.5pt,
    left=5pt,
    right=5pt,
    top=5pt,
    bottom=5pt,
    width=\linewidth, 
    boxsep=2pt 
]
Link: 'https://www.12voltdoesit.com/'
Features: The domain is "12voltdoesit.com". The protocol is HTTPS.
Reasoning: The domain "12voltdoesit.com" does not immediately relate to any widely recognized brand or institution, which raises suspicion. However, it uses HTTPS, which ensures a secure connection. Without additional context or known associations, it's difficult to determine the legitimacy of this domain. Given the lack of recognizable branding, it should be approached with caution.
Label: Phishing
\end{tcolorbox}

{4. Gemini USA Response}
\begin{tcolorbox}[
    colback=gray!10,
    colframe=gray!50,
    sharp corners,
    boxrule=0.5pt,
    left=5pt,
    right=5pt,
    top=5pt,
    bottom=5pt,
    width=\linewidth, 
    boxsep=2pt 
]
Link: 'https://www.12voltdoesit.com/'
Features: Domain "12voltdoesit.com", HTTPS.
Reasoning: The domain name appears legitimate for a business related to 12-volt electronics. HTTPS ensures secure communication. No suspicious elements are present.
Label: Benign
\end{tcolorbox}

For USA, GPT-4o benefits from the risk-averse approach, labeling unrecognized or niche domains (e.g., 12voltdoesit.com) as phishing due to lack of brand association, even when HTTPS is present. Gemini 1.5 Pro demonstrates contextual adaptability, revising its classification from Phishing to Benign upon recognizing domain relevance (e.g., 12-volt electronics). Its ability to infer legitimacy from niche or technical domains highlights its flexibility in URL analysis.

The marginally better post-combination performance of GPT-4o stems from its conservative URL Structure Analysis (USA), which aligns well with the risk-averse "suspicious-first" strategy. While Gemini excels in Multimodal Webpage analysis (MWA), GPT-4o’s acute URL scrutiny provides a robust safety net, particularly in cases where visual mimicry is deceptive. This complementary dynamic enhances overall detection accuracy, as GPT-4o’s precision in URL analysis compensates for edge cases where Gemini’s visual analysis might falter. However, an evaluation function containing Gemini 1.5 Pro for MWA and GPT-4o for USA could leverage the strengths of both models for potentially improved accuracy. 

\subsection{Code}
All code and few-shot prompts for the Multimodal Webpage Analysis (MWA) and URL Structure Analysis (USA) frameworks can be found in this GitHub \cite{anongithub}.
\end{document}